\documentclass[preprint,12pt]{elsarticle}

\usepackage{amssymb}
\usepackage{amsmath}
\usepackage{float}
\usepackage{listings}
\usepackage{amssymb}
\usepackage{tabularx}
\usepackage{tikz}
\usepackage{float}
\usepackage{mathtools}
\usepackage[colorinlistoftodos]{todonotes}
\usepackage[ruled,vlined]{algorithm2e}
\usepackage{hyperref}
\usepackage{enumerate}
\usepackage[shortlabels]{enumitem}
\usepackage{amssymb}
\usepackage{subfig}
\usepackage{makecell}
\usepackage{array}
\usepackage{multirow}
\usepackage{mathpazo}
\usepackage{siunitx}
\usepackage{graphicx}
\usepackage{longtable}
\usepackage[toc,page]{appendix}
\usepackage{booktabs}
\usepackage{tabularx}
\usepackage{adjustbox}
\usepackage[section]{placeins}
\usepackage{pdflscape}
\usepackage[utf8]{inputenc}
\usepackage[T1]{fontenc}
\usepackage{xcolor}

\usepackage{natbib}
\usepackage{todonotes}

\newcommand{\R}{\mathbb{R}}

\journal{Nuclear Instruments and Methods
in Physics Research Section A}

\begin{document}

\begin{frontmatter}


\title{Unsupervised Learning  for Identifying  Events in Active Target  Experiments}


\author{R.~Solli}
\address{Expert Analytics AS, Tordenskiolds gate 6, 0160, Oslo, Norway}
\address{Department of Physics, University of Oslo, POB 1048 Oslo, N-0316 Oslo, Norway}
\ead{robert@xal.no}
\author{D.~Bazin}
\address{Department of Physics and Astronomy and Facility for Rare Ion Beams and National Superconducting Cyclotron Facility, Michigan State University, East Lansing, MI 48824, USA}
\author{M.~Hjorth-Jensen}
\address{Department of Physics and Astronomy and Facility for Rare Ion Beams and National Superconducting Cyclotron Facility, Michigan State University, East Lansing, MI 48824, USA}
\address{Department of Physics and Center for Computing in Science Education, University of Oslo, POB 1048 Oslo, N-0316 Oslo, Norway}
\author{M.P.~Kuchera}
\address{Department of Physics, Davidson College, Davidson, North Carolina, USA}
\author{R.R.~Strauss}
\address{Department of Mathematics and Computer Science, Davidson College, Davidson, North Carolina, USA}
\address{Department of Computer Science, University of North Carolina, Chapel Hill, North Carolina, USA}

\begin{abstract}
This article presents novel applications of unsupervised machine learning methods to the problem of event separation in an active target detector, the Active-Target Time Projection Chamber (AT-TPC) \cite{Bradt2017}. 
The overarching goal is to group similar events in the early stages of the data analysis, thereby improving efficiency by limiting the computationally expensive processing of unnecessary events.
The application of unsupervised clustering algorithms to the analysis of two-dimensional projections of particle tracks from a resonant proton scattering experiment on $^{46}$Ar is introduced. We explore the performance of autoencoder neural networks and a pre-trained VGG16 \cite{Simonyan2014} convolutional neural network. We study clustering performance on both data from a simulated $^{46}$Ar experiment, and real events from the AT-TPC detector. 
We find that a $k$-means algorithm applied to simulated data in the VGG16 latent space forms almost perfect clusters. Additionally, the VGG16+$k$-means approach finds high purity clusters of proton events for real experimental data. We also explore the application of clustering the latent space of autoencoder neural networks for event separation. While these networks show strong performance, they suffer from high variability in their results.  

\end{abstract}




\end{frontmatter}

\section{Introduction}\label{sec:intro}

The Active-Target Time Projection Chamber (AT-TPC) \cite{Bradt2017} is a novel type of detector designed specifically for nuclear physics experiments where the energies of the recoiling particles are very low compared to the energy required to escape the target material. The luminosity of nuclear physics experiments performed with fixed targets is directly proportional to the amount of material encountered by the beam. On the other hand, for several classes of experiments the detection of recoil particles is paramount, therefore limiting the target thickness. In addition, the properties of the recoil particles are modified while traversing the target material, affecting the resolutions that can be achieved. This necessary balance between luminosity and resolution is particularly difficult when performing experiments with rare isotope beams, chiefly because of the low intensities available. 

The concept of active targets aim at mitigating this compromise, by turning the target itself into a detector \cite{BECEIRONOVO2015}. Most active target detectors such as the AT-TPC are composed of a time projection chamber (TPC) where the detector gas is at the same time the target material. Recoil particles that originate from a nuclear reaction between a beam nucleus and a gas nucleus can be tracked from the vertex of the reaction to their final position inside the active volume of the target. Their properties can therefore be measured without any loss of resolution regardless of the amount of material traversed by the beam. At the same time, the detection efficiency is dramatically increased by the very large solid angle covered in this geometry. A direct consequence of this concept is the inclusiveness of the experimental data recorded by this type of detector: any nuclear reaction happening within the target is recorded. Although this sounds like an advantage from the scientific point of view, it poses great challenges during the analysis phase that are reminiscent 
of bubble-chamber times and on par with event classification challenges in particle physics today, see for example the recent review of Mehta {\em et al.} \cite{mehta2019}. More often than not, the reaction channel of interest has one of the lowest cross sections. When analyzing the data one is therefore faced with the task of sorting out the corresponding events from the {\em background} of other reaction channels. 

Because TPCs produce three-dimensional images of charged particle tracks, the event identification task is often akin to a visual inspection (comparable to the analyses made in the bubble-chamber era \cite{Hough:1959}), which is not practical nowadays because of the large quantities of data. Machine learning techniques then appear as a promising prospect, in particular in the image recognition domain where much progress has been made recently \cite{mehta2019}. In addition, machine learning (ML) algorithms offer new possibilities such as the potential discovery of unforeseen phenomena that would have been missed by more traditional analysis methods. Prior work has demonstrated the ability to apply supervised classification machine learning methods to AT-TPC data when a labeled training set is available, whether through hand-labeled data or labeled simulated data (a {\em transfer learning} application) \cite{Kuchera2019}. In some experiments, a labeled data set is unavailable. This could be  due to the inability to hand-label events, or the case where one does not know {\em a priori} the types and behaviors of the reactions present in the detector in order to generate a labeled and simulated data set. In the latter case, there must also exist a validation data set of real data, still requiring the ability to label a subset of the real data. The unsupervised separation of event types, or {\em clustering}, based on a set of ML algorithms is hereby examined, using experimental data recorded by the AT-TPC during its commissioning experiment from a radioactive $^{46}$Ar beam reacting on an isobutane target composed of proton and carbon nuclei. 

Within the context of machine learning methods applied to the analysis of nuclear physics experiments, 
the purpose of this work is thus to explore the application of unsupervised learning algorithms to event identification from an active target detector. The necessity to identify events from raw data prior to full processing is becoming a major issue in the data analysis of detectors with complex responses such as the AT-TPC.  After these introductory words, we review the experimental information in the next section. In section \ref{sec:datadescript} we describe the set up of the data while sections \ref{sec:methods} and \ref{sec:results}
present the methods applied and our results and discussions, respectively. Finally, our conclusions and perspectives for future work are presented in section \ref{sec:conclusion}.

\section{Experimental Details} 
\label{Sec:Exp}
The goal of the experiment was to measure the evolution of the elastic and inelastic cross sections between $^{46}$Ar and protons as a function of energy (the {\em excitation function}), and observe resonances in the composite system $^{47}$K that correspond to analog states in the nucleus $^{47}$Ar. Spectroscopic information can then be obtained from the shape and amplitude of the observed resonances \cite{Bradt2018}. 
The experiment was performed at the National Superconducting Cyclotron Facility (NSCL) where a $^{46}$Ar beam was produced via fragmentation of a $^{48}$Ca beam on a $^9$Be target at about 140 MeV/u. The $^{46}$Ar isotopes were then filtered, thermalized, and finally re-accelerated to 4.6 MeV/u by a linear accelerator. This scheme was used to produce a low-emittance beam, which is necessary to guarantee a good energy resolution in the excitation function. Because the  $^{46}$Ar beam particles lose energy as they traverse the target gas volume, the position of the reaction vertex along the beam axis is directly related to the energy at which the reaction occurs. This allows the AT-TPC to measure the excitation function over a wide range of energies from a single beam energy.

The detector was placed inside the bore of an MRI solenoid energized to $\sim 2$ Tesla. This axial magnetic field served the purpose of bending the trajectories of the recoil particles in order to i) increase their length and ii) provide a measurement of their bending radius, directly related to their magnetic rigidity. Because the recoil particles travel in gas, they slow down and eventually stop, therefore their trajectories are described by three-dimensional spirals (see \cite{Bradt2017}). One of the difficulties encountered in the analysis is that the shape of these spirals does not have an analytical form because it follows the energy-loss profile of the particles. It therefore needs to be simulated via an integration, which is numerically costly. Common integration methods such as Runge-Kutta have a computational cost typically one of magnitude larger than an analytical calculation. Other difficulties are related to several experimental effects that deteriorate the quality of the data, namely saturation and cross-talk effects, as well as random noise. 

The method used in \cite{Bradt2018} to analyze the data followed a three-phase sequence: cleaning, filtering and fitting. Traditional methods were used to perform each of these tasks, and ultimately extract the scientific information, but there were severe limitations and high computational costs that become prohibitive in data sets larger by an order of magnitude.  In particular, as for the data presented and analyzed here, with data sets in the terabyte region or larger, these more traditional methods become prohibitively expensive from a computational stand. This is particularly the case when using the Monte-Carlo fitting procedure (explained later in text), because of the thousands of calculated tracks that are required for each individual event. For instance, the analysis of the $^{46}$Ar(p,p) data required several seconds per event on a CPU cluster composed of about 100 nodes.

The cleaning was performed using a combination of linear and circular Hough transforms on a 2-dimensional projection of the tracks \cite{Bradt2017}. The following filtering and fitting phases were performed simultaneously, by defining the cost function to the fitting algorithm as a sum of three $\chi^2$ components based on i) the position of the track in space, ii) the energy deposited on each pixel of the sensor plane, and iii) the location of the vertex of the reaction. While various fitting algorithms were tested, the most accurate was a Monte-Carlo algorithm that explored the six-dimensional phase space of the particle's kinematics parameters, reducing it progressively at each iteration step until the desired accuracy was reached \cite{Bradt2017}. Although this algorithm ended up being the most accurate, it is extremely costly computationally because of the very large number of simulated tracks needed for each event. The filtering was performed by setting limits to the $\chi^2$ distributions, below which the events were assigned as proton scattering. This is a very inefficient method because it requires performing the fitting for all events, including those that are not of interest. Pioneering work on event identification using machine learning methods, namely the use of a pre-trained convolutional neural networks (CNN), later showed that the filtering phase would better be performed using this type of technique \cite{Kuchera2019}. 
In addition, the authors of Ref.~\cite{Kuchera2019} demonstrated that the purity and statistics of the data are improved with the use of machine learning methods.

From the experimenter's point of view, it is clear that the method used in \cite{Bradt2018} to identify and filter events is not the most efficient computationally. The ML methods explored in \cite{Kuchera2019} are a step forward, but they still rely on supervised learning methods that require data labeling, a time-consuming and error-prone process. The aim of the present study is to investigate unsupervised learning methods that bypass the labeling step, and form classes of events independently from the experimenter's input. The task of labeling the different classes is then much less time-consuming and can potentially lead to the discovery of unforeseen types of events. Furthermore, it allows us to process larger amounts of data in a much shorter time. 

\section{Data Preparation}\label{sec:datadescript}

In this section we give a brief overview of the data, for a more in-depth consideration we refer the reader to Refs.~\cite{Mittig2015,Suzuki2012,Bradt2017a}. 

The AT-TPC data we studied for this work was recorded as charge time-series for each of the  the $\sim10^4$ detector pads.
In this representation, an event is a record of $512$ time-buckets for each of the $10^4$ detector pads. In our analysis, we represent each event as a down-sampled two-dimensional  projection.
We chose to represent the data in two dimensions to facilitate the use of advanced image-recognition machine learning models, and this data representation was shown to successfully classify the events of this experiment in a supervised setting \cite{Kuchera2019}.
First, the time-series data was represented as a three-dimensional  point cloud, where each point contains the maximum charge in the time-series trace. We log-transform the data and perform a min-max scaling in order to  map the data to the interval $[0, 1]$. The data is projected onto a two-dimensional  space by summing over the time-axis. The two-dimensional data is then down-sampled into a $128\times128$ pixel image by discretizing the space in the $x-y$ plane and summing all charge values that fall within the bounds for each element location.

One of the significant considerations for the analysis of AT-TPC data is to inject machine learning methods for track identification at the best point in the analysis pipeline.
Using raw data is advantageous as it provides an unbiased view of the event, but the data volume and noise levels might be prohibitive for the analysis.
Therefore, we incrementally add bias to the analysis by applying the algorithm further down the analysis pipeline, with the benefit being that more preprocessing improves the signal-to-noise ratio, possibly improving model performance.
To explore this trade-off between model performance and preprocessing bias, we performed our analysis on simulated, raw and cleaned events as discussed below.



\subsection{Simulated \texorpdfstring{${}^{46}$Ar}{46Ar} events}\label{sec:data_sim}

A set of $N=8000$ simulated AT-TPC events per class was generated from the \texttt{pytpc} package developed by \citet{Bradt2017a} for the analysis of the  ${}^{46}$Ar$(p, p)$ experiment.

For validation, we select a subset of the simulated data to be labeled and treat the rest as unlabeled data. We chose this partition to consist of  $15\%$ of the data. We denote this subset and its associated labels as $\gamma_L=(\boldsymbol{X}_L, \boldsymbol{y}_L)$, while the entire data set is identified as $\boldsymbol{X}_F$. Note that $\boldsymbol{X}_L \subset \boldsymbol{X}_F$.
As this dataset is generated via simulation, the true labels are known for this dataset.

\subsection{Raw \texorpdfstring{${}^{46}$Ar}{46Ar}  events}\label{sec:data_real}

The events analyzed in this section were retrieved from the ${}^{46}$Ar resonant proton scattering experiment recorded with the AT-TPC. 
While we denote these events as raw, it is important to note that what we intend is a raw two-dimensional projection of events without any post-processing done to clean the data.

\begin{figure}[ht]
\centering
\includegraphics[width=\textwidth]{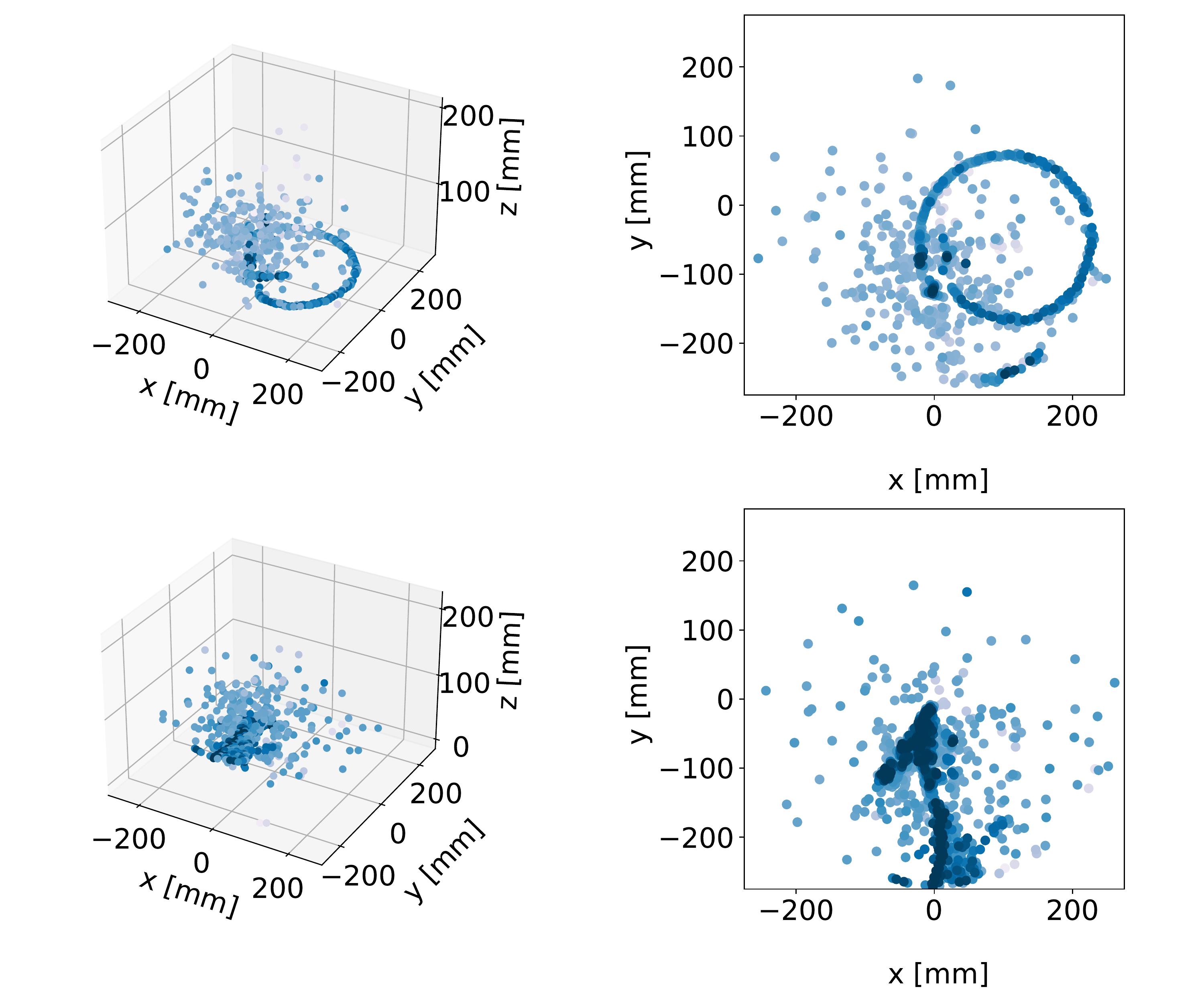}
\caption[Displaying unfiltered events in two and three dimensions]{Two- and three-dimensional representations of two events from the ${}^{46}$Ar experiment. Each row is one event in two projections, where the color intensity of each point indicates higher charge values recorded by the detector. The top row illustrates a proton event with a distinctive spiral-shaped track, while the bottom row shows a carbon event with a large fraction of noise. The beam direction is oriented along the z-axis (from positive to negative), close to the (x,y) origin. The magnetic field is oriented on the same axis as the beam, causing the bending of the charged particle's trajectories. The spiral shape is caused by the slowing down of the particles inside the gas volume. The bending radius of the tracks depends not only on the energy, but also on the mass-to-charge ratio of the particles, hence the large difference between proton and carbon events (see also section \ref{Sec:Exp}).}
\label{fig:samples}
\end{figure}

We display two different events from the ${}^{46}$Ar experiment in Fig.~\ref{fig:samples}. The top row illustrates a proton event with a distinctive spiral-shaped track, while the bottom row shows a carbon event with a large fraction of noise. 
A subset of this data was hand-labeled for validation, as discussed in \cite{Kuchera2019}.

\subsection{Filtered \texorpdfstring{${}^{46}$Ar}{46Ar} events}\label{sec:filtered}

As we saw in the previous section, the detector picks up a significant amount of noise. We split the noise broadly in two categories,  one being randomly uncorrelated noise and the second one being structured noise. The former can be quite trivially removed with a nearest-neighbor algorithm, see for example \cite{hastie2009}, that checks if a point in the event is close to any other. To remove the correlated noise, researchers at the National Superconducting Cyclotron Laboratory of Michigan State University, developed an algorithm based on the Hough transform \cite{Newman1972}. This transformation is a common technique in computer vision, used to identify common geometric shapes like lines and circles, and has been used extensively in high-energy particle physics since the bubble-chamber era \cite{Hough:1959}.  

\begin{figure}[ht]
\centering
\includegraphics[width=0.9\textwidth, height=9cm]{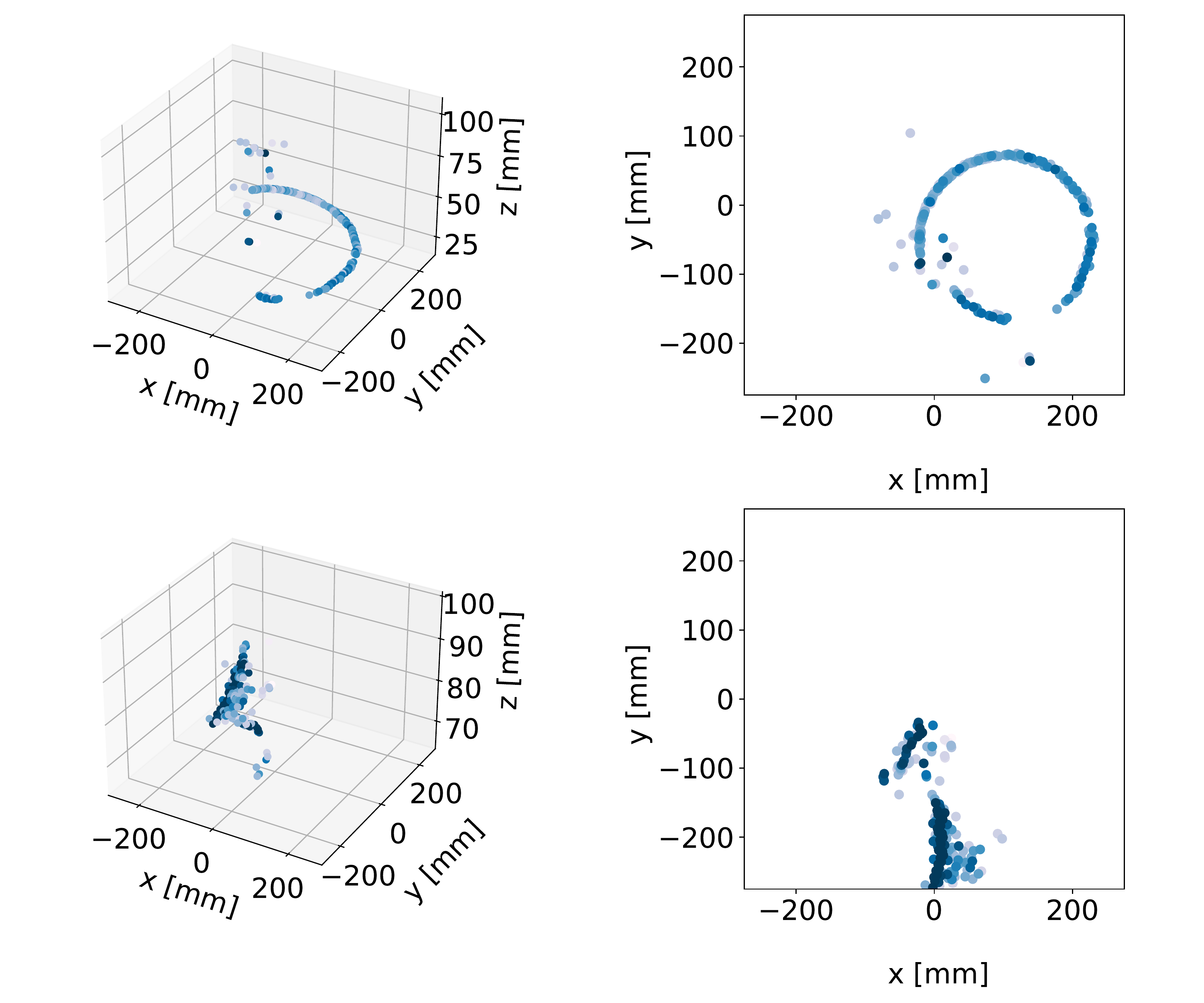}
\caption[Displaying filtered events in 2D and 3D]{Two- and three-dimensional representations of two events from the ${}^{46}$Ar experiment. Each row is one event in two projections, where the lightness of each point indicates higher charge values. These events have been filtered with a nearest neighbors algorithm and an algorithm based on the Hough transform \cite{Newman1972}.}\label{fig:samples_filtered}
\end{figure}

We illustrate two filtered events in Fig.~\ref{fig:samples_filtered}. These are the same events as shown in Fig.~\ref{fig:samples}, but with the Hough transform and nearest-neighbor filtering applied. 
The same subset of hand-labeled data from Section~\ref{sec:data_real} was used for validation.

\section{Methods}\label{sec:methods}

\subsection{Classifying  events} 
The traditional Monte Carlo event selection process, described in Section~\ref{Sec:Exp}, does not have a well-defined method to quantify the effectiveness of the event selection.
In addition, the selection task produced a binary, cut-based result: either a {\em good} or {\em bad} fit to the event of interest. A {\em bad} fit is then assumed to be a different event type, and is removed from the analysis. 
In a broader perspective, an unsupervised classification algorithm would offer the possibility to {\em discover} rare events which may not be expected or are overlooked. These events would likely be filtered out using the traditional methods. From a practical point of view, compared to supervised learning, it also avoids the necessary labeling task of the learning set events, which is error prone and time consuming.

\subsection{Why machine learning}

The $\chi^2$ approach used in the traditional analysis performed on the $^{46}$Ar data is extremely expensive computationally because it requires the simulation of thousands of tracks for each recorded event. These events are in turn simulated for each iteration of the Monte Carlo fitting sequence. Even though the reaction of interest in the $^{46}$Ar experiment had the largest cross section (elastic scattering), the time spent on Monte Carlo fitting of {\em all} of the events produced in the experiment was the largest computational bottleneck in the analysis. In the case of an experiment where the reaction of interest would represent less than a few percent of the total cross section, this procedure would become highly inefficient and prohibitive. Adding to this the large amount of data produced in this experiment (with even larger data sets expected in future experiments), the analysis simply begs for more efficient analysis tools. 
The computationally expensive fitting procedure would be applied to every event, instead of the few percent of the events that are of interest for the analysis.
An unsupervised ML algorithm able to separate the data without {\em a priori} knowledge of the different types of events increases the efficiency of the analysis tremendously, and allows the downstream analysis to concentrate on the fitting efforts only on events of interest. In addition, the clustering allows for more exploration of the data, potentially enabling new discovery of unexpected reaction types.





\subsection{Pre-trained neural networks}

Training high-performing neural networks from scratch often requires enormous data sets and computation time. However, it has been found that models which are trained at large scale will learn general features that are applicable to a variety of tasks. For example, large neural networks which are trained on the ImageNet data set \cite{deng2009imagenet} --- a diverse image classification task --- learn how to identify lines, edges, and other common shapes that are useful for numerous problems. Thus, it is common practice to initialize the convolutional layers of a network with the pre-trained weights learned from ImageNet (or some other large data set). The training process then only has to fine-tune the network for the specific task. Since we are building on prior knowledge in this case, learning becomes far more efficient, and better performance can often be achieved. \citet{Kuchera2019} used machine learning methods to classify the products of $^{46}$Ar reactions in the AT-TPC, and they found that a CNN initialized with weights trained on ImageNet data resulted in the most successful classification.

\subsection{Clustering on latent spaces}



In contrast with the classification work of \citet{Kuchera2019}, we do not assume access to ground truth labels and we are trying to solve a fundamentally different learning problem. Thus, rather than fine-tuning a pre-trained network under the supervised learning regime, we extract the output of the pre-trained network's last convolutional layer as a latent representation of the events, where each event is represented as a vector in $\R^{8192}$. We then cluster events based on this representation using the \texttt{scikit-learn} implementation of the $k$-means algorithm with default parameters \cite{Pedregosa2011}.

\subsection{Deep clustering: Mixture of autoencoders}\label{sec:mixae}

As an alternative to relying on a pre-trained model, we also consider the MIXAE algorithm \cite{Zhang}, which is an end-to-end clustering model specifically trained on the AT-TPC data.

The MIXAE model comprises several autoencoders, each of which corresponds to a cluster.  Each autoencoder constructs a latent representation of a given example. Those representations are used as inputs to an auxiliary network which assigns scores to the clusters, indicating the likelihood that the given example belongs to each cluster. Examples are then assigned to the cluster with the highest score.  
The number of autoencoders, and thus the number of clusters, has to be determined beforehand.

The MIXAE algorithm relies on a few simple assumptions which are necessary, but not sufficient, for producing a high-quality model. The assumptions can be stated as: 
\begin{enumerate}
	\item  If an example is assigned to a particular cluster, the corresponding autoencoder's reconstruction should be accurate. 
	\item Within a batch of examples presented to the model, assignments should spread across all clusters. 
	\item Each clustering prediction should be as strong as possible, i.e. assigning high probabilities is preferable to weak assignments.
\end{enumerate}

The learning objective encourages these assumptions to be met. For a more formal consideration of the model objective and the assumptions made on the data by this model see \citet{Zhang}.

\begin{figure}[tbh]
	\centering
	\includegraphics[width=.8\textwidth]{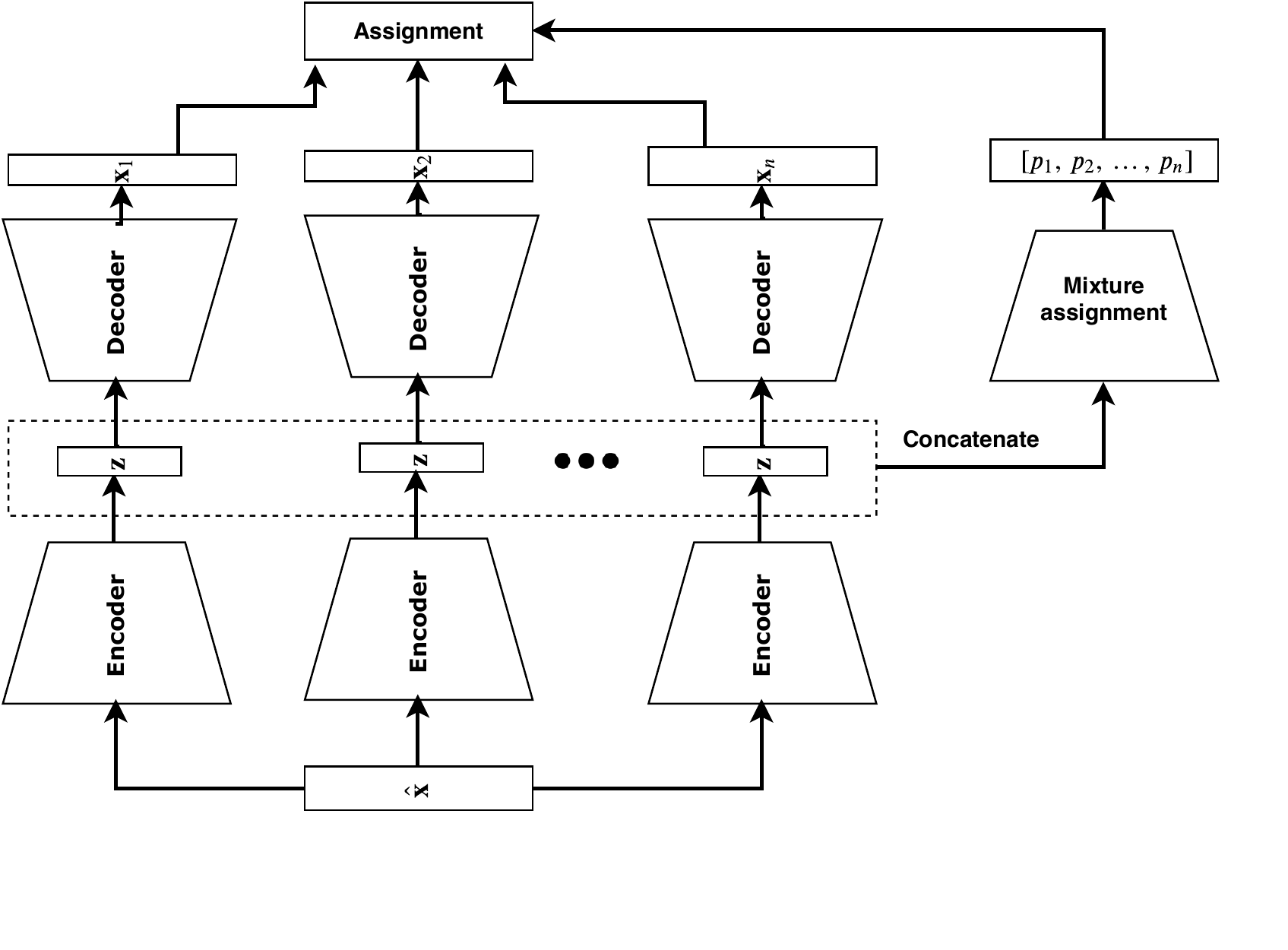}
	\caption[Mixture of autoencoders schematic]{Schematic of a MIXAE model. A sample $\hat{\boldsymbol{x}}$ is compressed to set of lower-dimensional representations $\{\boldsymbol{z}^{(i)}\}$ by $N$ autoencoders \cite{Zhang}. These samples are concatenated and passed through an auxiliary assignment network that predicts a confidence of cluster belonging for each autoencoder.}
	\label{fig:mixae}
\end{figure}

The architecture is portrayed in Fig.~\ref{fig:mixae}, wherein tapered boxes denote a direction of compression in the network components. In the figure each encoder and decoder pair makes up an autoencoder. The assignment of the cluster for a sample x-y event image, $\mathbf{\hat{x}} \in \mathcal{R}^{128 \times \hspace{0.01in} 128} $, is taken to be the index of the maximal element in the vector $\mathbf{p}$ as shown in the right-most part of the figure. Finally, the model is trained end-to-end with back propagation \cite{Linnainmaa1976}, as implemented in the machine learning package TensorFlow \cite{tensorflow}.

\subsection{Measuring performance}
Unsupervised machine learning  is often accompanied with a  lack of ground-truth labelled data. In the face of such missing data, a model is ordinarily assessed by measures that do not depend on knowing the ground truth for any samples. 
However, as our work aims to explore the application of unsupervised methods to track identification, we chose the ${}^{46}$Ar data since we have some ground truth labels to evaluate our models. The measures we use to evaluate the models we then chose to be those which measure the similarities between two arbitrary sets of clustering assignments, while holding one to be the ground truth.

We measure the performance of the clustering algorithms by two functions: the clustering accuracy and the adjusted rand index (ARI)\cite{Hubert1985}. Both of these measurements fall in the range $ [0, \hspace{0.1cm}1]$, where $0$ denotes a complete disagreement and $1$ a complete agreement between the ground truth and predicted labels.

To compute the metrics we have to solve problems introduced by the arbitrary labels of a clustering algorithm. That is, we do not know which predicted assignment should correspond to a proton or carbon event. In short, we want to find the most reasonable correspondence between the clusters and the ground truth classes. To solve this problem we first define two sets; the ground truth labels $\mathbf{y} = [y_i, \hspace{0.2cm} y_{i +1}, \ldots]$ as determined in section \ref{sec:datadescript}, and the corresponding predictions $\mathbf{\hat{y}} = [\hat{y}_i, \hspace{0.2cm} \hat{y}_{i +1}, \ldots]$. Furthermore, we let both $y_i$ and $\hat{y}_i$ be integer representations of an event's ground truth, and predicted class, respectively. 

To compute both the ARI and the clustering accuracy we also have to construct the contingency table between two sets of clustering assignments. A contingency matrix defines the overlaps between classes in these sets, and its general form is shown in Table \ref{tab:contingency}.

\begin{table}[H]
\centering
\caption{General form of a contingency table. Here $y_n$ and $\hat{y}_n$ are the ground truth labels and clustering assignments respectively. The $w_{ij}$'s then describe how many samples are in clusters $y_i$ and simultaneously in $\hat{y}_j$.}\label{tab:contingency}
\begin{tabular}{c|ccc|c}
& $y_1$ &$y_2$ & $\cdots$ & sums \\
\midrule
$\hat{y}_1$ & $w_{11}$ & $w_{12}$ &  $\cdots$ & $e_1$ \\
$\hat{y}_2$ & $w_{21}$ & $w_{22}$ &  $\cdots$ & $e_2$\\
$\vdots$ & $\vdots$ & \vdots & $\ddots$ & \vdots \\
\hline
sums & $f_1$ & $f_2$ & $\hdots$ 
\end{tabular}
\end{table}

The algorithm for finding the clustering accuracy can then be described in these steps: 

\begin{itemize}
    \item Compute the matrix $\mathbf{W}$ from the contingency table between $\mathbf{y}$ and $\mathbf{\hat{y}}$
    \item Subtract $\mathbf{W}$ from its maximum value to find the bipartite graph representation of the assignment problem.
    \item Use an algorithm, like the Hungarian algorithm \cite{burkard2012}, to solve the assignment problem, and take the average of the values in $\mathbf{W}$ this solution indicates. This average is the clustering accuracy. 
\end{itemize}

To compute the ARI we use the elements from the contingency table to evaluate the function introduced by \citet{Hubert1985}

\begin{equation}\label{eq:ari}
\text{ARI} = \frac{\sum  \binom{w_{ij}}{2} - \left[\sum  \binom{e_{i}}{2} \sum  \binom{f_{j}}{2}  \right]/\binom{n}{2}}{\frac{1}{2}\left[\sum  \binom{e_{i}}{2} + \sum  \binom{f_{j}}{2}  \right]- \left[\sum  \binom{e_{i}}{2} \sum  \binom{f_{j}}{2}  \right]/\binom{n}{2}}.
\end{equation}
The quantities $f_i$ and $e_i$ are defined in Table \ref{tab:contingency}.
The important distinction to make between the clustering accuracy and the ARI is that the clustering accuracy is a simple comparison that does not account for chance assignments, while the ARI does. In effect, this means that the accuracy is a good heuristic for the performance, tempered by the ARI. 

Lastly, we introduce two terms to describe cluster quality: purity and quality. Purity is inferred by how much spread there is in the column between the ground truth labels in the matrix $\mathbf{W}$. A high-quality cluster will, in addition to being pure, also capture most entries the class represented by the cluster.

The performance is measured by comparing model predictions on the labelled subset of data (see Table \ref{tab:data sets} in  appendix B for more details).

\section{Results and Discussions}\label{sec:results}

The principal challenge in the AT-TPC experiments that we are trying to address is the reliance on labelled samples in the analysis, as future experiments may not have as visually dissimilar reaction products as we observe in the ${}^{46}$Ar experiment.  The ability to label data in the the ${}^{46}$Ar experiment does, however, provide a useful example where we can then explore unsupervised techniques. 

We first explore the results of applying a $k$-means approach on the latent space of a pre-trained network. Subsequently, we investigate the performance of the MIXAE algorithm as outlined in section \ref{sec:mixae}.

\subsection{$k$-means clustering on the VGG16 latent space}\label{sec:kmeans_results}

The results of the clustering runs are included in Table \ref{tab:clstr_vgg1}. We ran the $k$-means algorithm $N=100$ times with $M=10$ initializations per run, of the cluster centroids to assess the variability in the performance. The $k$-means algorithm returns the best performing model of the $M$ initializations on the unsupervised objective. We report performance on the labelled subset of data, using the labels to identify the top-performing model (Top 1). Additionally, we report the mean and standard deviation of the algorithm on the $N$ trials, which indicate unsupervised performance.   


We observe that the clustering on simulated data attains the highest performance, and that there is a decline in performance as we add noise by moving to the filtered and raw data sets. The results are shown in Table \ref{tab:clstr_vgg10}. 

\begin{table}[H]
\centering 
\caption[$k$-means on pre-trained model]{$k$-means clustering results on AT-TPC event data in the VGG-16 latent space, for $N=100$ runs of the $k$-means algorithm with $M=10$ initializations. We observe that the performance predictably decreases with the amount of noise in the data.}\label{tab:clstr_vgg10}
\begin{tabular}{lllll}
\toprule
{} & \multicolumn{2}{c}{Accuracy} &   \multicolumn{2}{c}{ARI} \\
\midrule
{} & Top 1 & $\mu \pm \sigma$ &  Top 1 & $\mu \pm \sigma$ \\
Simulated &  $0.97$ & $0.97 \pm 0.0 $ &  0.89 & $0.89 \pm 0.0 $ \\
Filtered  &   $0.75$  & $0.75 \pm 0.0$ &  0.40 & $0.40 \pm 0.0$ \\
Raw &   $0.59$ &  $0.59\pm 0.0$ & $0.17$ & $0.17\pm 0.0$ \\
\bottomrule
\end{tabular}

\end{table}

\noindent The lack of variability is explained by the number of initializations. As can be seen from Table \ref{tab:clstr_vgg1} where we run the $k$-means algorithm $N=1000$ times with $M=1$ initializations of the centroids. 

\begin{table}[H]
\centering 
\caption[$k$-means on pre-trained model]{$k$-means clustering results on AT-TPC event data in the VGG-16 latent space, for $N=1000$ runs of the $k$-means algorithm with $M=1$ initializations. We observe that there is significant variability in the results, which is ordinarily masked by $M$ re-initializations that avoid local minima. }\label{tab:clstr_vgg1}
\begin{tabular}{lllll}
\toprule
{} & \multicolumn{2}{c}{Accuracy} &   \multicolumn{2}{c}{ARI} \\
\midrule
{} & Top 1 & $\mu \pm \sigma$ &  Top 1 & $\mu \pm \sigma$ \\
Simulated &  $0.97$&  $0.86 \pm 0.18 $ &  0.89 & $0.63 \pm 0.39 $ \\
Filtered  &   $0.75$  & $0.75 \pm 0.0$ &  0.40 & $0.40 \pm 0.0$ \\
Raw &   $0.71$ &  $0.59\pm 0.019$ & 0.29 & $0.18\pm 0.018$ \\
\bottomrule
\end{tabular}

\end{table}

In addition to the performance measures reported in Table \ref{tab:clstr_vgg1}, it is interesting to observe which samples are wrongly assigned. To investigate this problem, we compute the matrices $\mathbf{W}$ as shown in Table \ref{tab:contingency}. From these tables, we can infer which classes are more or less entangled with others. The results for each data set is shown in  Figs.~\ref{fig:clster_confmat_sim}, \ref{fig:clster_confmati_filt} and \ref{fig:clster_confmati_raw}. We observe that the proton class is consistently assigned in a pure cluster. For example, consider the row corresponding to the proton class in Fig.~\ref{fig:clster_confmati_filt}. The column corresponding to the largest entry in the proton row has zero other predicted classes in it.

This high-quality cluster also appears in the clustering of raw data. From Fig.~\ref{fig:clster_confmati_raw}, we observe that there is a high purity proton cluster. In contrast to the filtered data we observe that the deterioration in performance can largely be ascribed to the algorithm creating a proton plus another cluster and a carbon plus another cluster.

We repeat this analysis using a Principle Component Analysis (PCA) dimensionality reduction\footnote{
PCA is a common technique to find the significant variations in data by projecting the data along a subset of its covariance matrix eigenvectors \cite{vidal2016,Marsland2009}
} on the latent space of the VGG16 model. This is done to estimate to what degree the class separating information is encoded in the entirety of the latent space, or in some select regions. The results from the PCA analysis, using the top 100 principal components, were virtually identical to our previous results not containing  the PCA analysis. This in an interesting observation which indicates that the class-encoding information is contained in a minority of the axes of variation in the data.

\begin{figure}
\includegraphics[width=\textwidth]{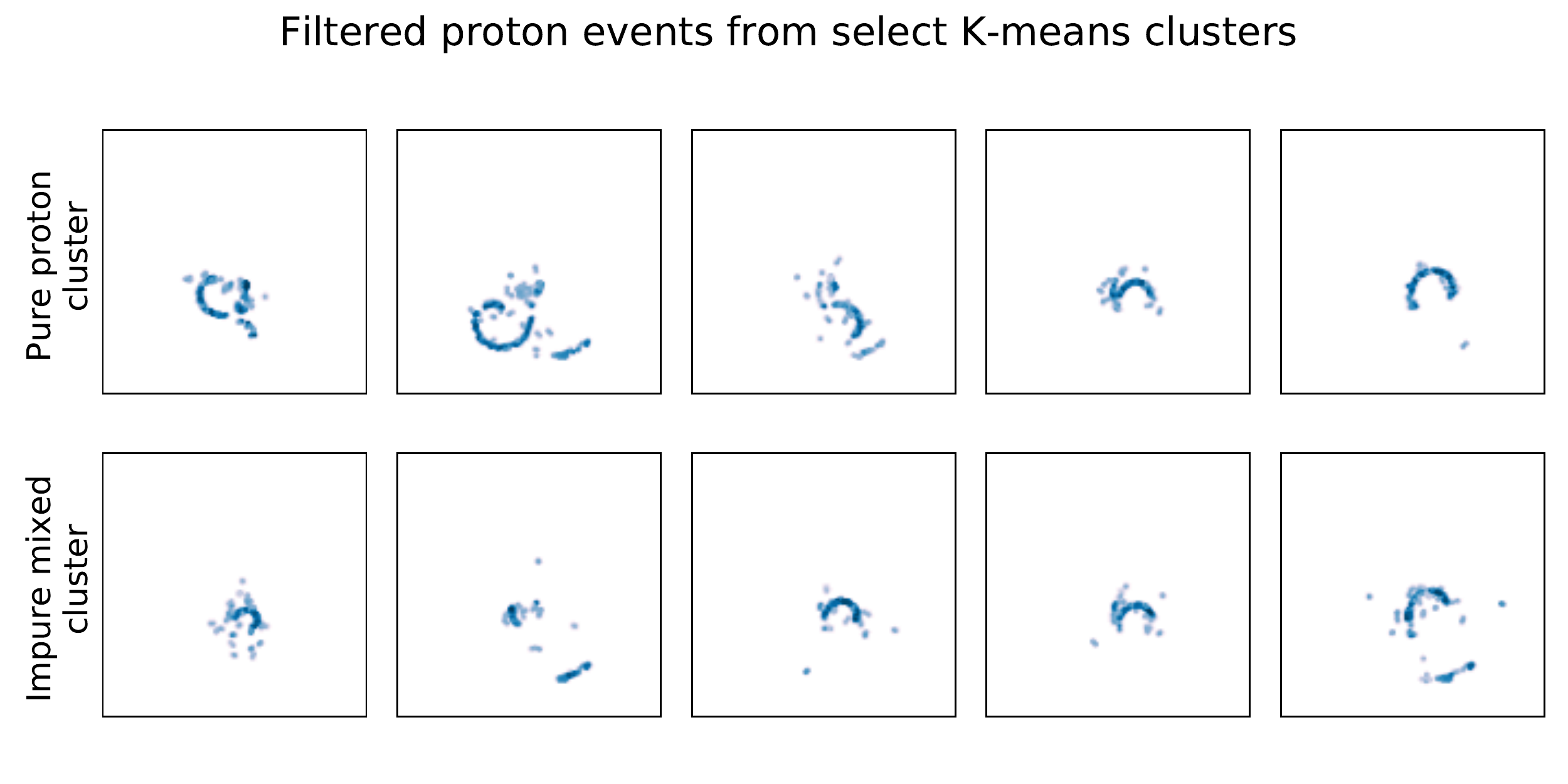}
\caption[Filtered proton samples by cluster belonging]{A sample of proton events from different $k$-means clusters from the filtered data set. The bottom row shows proton samples that are intermingled with noise events in the models' predictions, and the top row belongs to a cluster that contains proton events almost exclusively. Each row belongs to a single cluster corresponding to the filtered confusion matrix in Fig.~\ref{fig:clster_confmati_filt}. While the events in the pure proton cluster are visually distinct from the impure cluster, the difference is less pronounced than what we observe for the full data-set in figure \ref{fig:full_vgg_clster_repr}. }\label{fig:filtered_vgg_clster_repr}.
\end{figure}

\begin{figure}
\includegraphics[width=\textwidth]{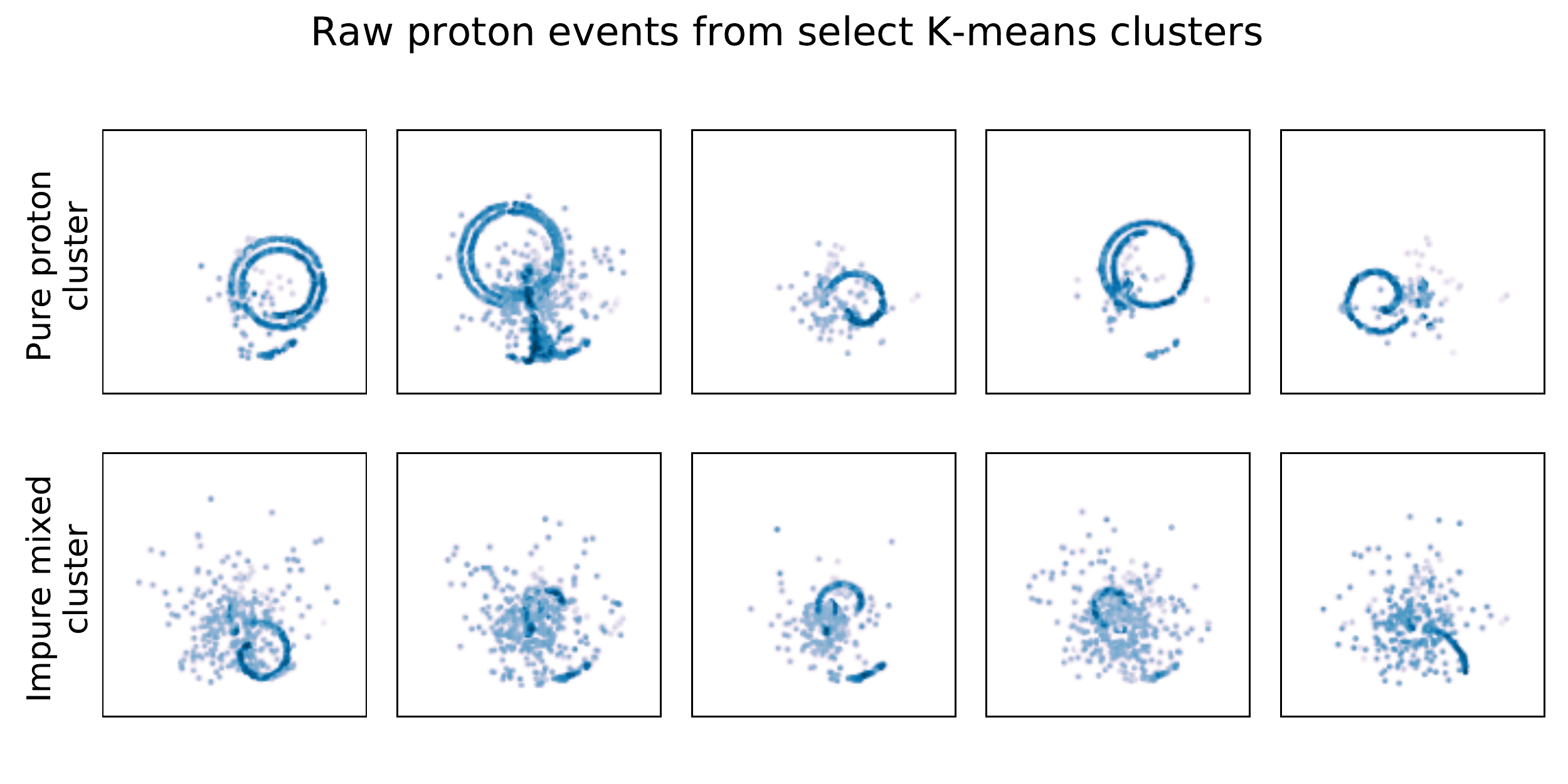}
\caption[Full proton samples by cluster belonging]{A random sample of proton events from different $k$-means clusters from the raw data set.  Each row belongs to a single cluster corresponding to the full confusion matrix in Fig.~\ref{fig:clster_confmati_raw}. The proton events in the cluster corresponding to the first row appear more visually distinct from other reaction types while the proton events corresponding to the cluster in the bottom row appear visually more similar to other reaction types in our dataset. }\label{fig:full_vgg_clster_repr}
\end{figure}

To further investigate the clusters presented in the matrix in Figs.~\ref{fig:clster_confmati_filt} and \ref{fig:clster_confmati_raw}, we visualize a random subset of examples from the proton events belonging to different clusters for the filtered and full data in Figs.~\ref{fig:filtered_vgg_clster_repr} and \ref{fig:full_vgg_clster_repr}, respectively.
We look at random samples of proton events in two clusters to infer an intuition on the track properties that are considered similar.
The figures indicate that shorter tracks, therefore low-energy or small scattering angle protons, are more likely to appear similar to other event types.

\subsection{MIXAE clustering results}

In the previous section we demonstrated a powerful, but rather naive, clustering technique for AT-TPC track identification. To build on this result we will in this section explore the application of the mixture of autoencoders (MIXAE) algorithm introduced in section \ref{sec:mixae}.
For details on hyper-parameter tuning and the experimental procedure for training the MIXAE algorithm see the \ref{ax: mixae}. 

With the best set of hyperparameters, each highest performing model is thereafter run $10$ times\footnote{The MIXAE model is significantly more computationally expensive to train than a $k$-means model. Resulting in the skew in number of runs in the two cases.}. The results are listed in Table \ref{tab:clstr_mixae}.
We observe that, while the algorithm can achieve a very strong performance, the performance varies. In some cases the MIXAE model converges to a seemingly good configuration, based on its unsupervised training goals. However, when inspecting its clustering performance against labelled data, the seemingly good model does no better than a model based on random selection. 
This happens more frequently with the raw data, indicating an interaction with the noise levels present in the events. 
 
\begin{table}[H]
\centering 
\caption[MIXAE clustering performance]{MIXAE clustering performance on the ${}^{46} Ar$ experimental data with $N=10$ runs of the algorithm. To quantify the results we report the best performing model (Top 1), and the mean and standard deviation for the result ($\mu\pm \sigma$). In contrast with the VGG-16 + $k$-means approach we observe  significant variations in performance.}\label{tab:clstr_mixae}
\begin{tabular}{lllll}
\toprule
{} & \multicolumn{2}{c}{Accuracy} &   \multicolumn{2}{c}{ARI} \\
\midrule
{} & Top 1 &  $\mu \pm \sigma$ & Top 1 & $\mu \pm \sigma$ \\
Simulated &  0.96 &  $0.74 \pm 0.16 $  & 0.84 &  $0.33 \pm 0.32 $\\
Filtered  & 0.75 &  $0.71 \pm 0.04 $ & 0.52 &  $0.38 \pm 0.14 $\\
Raw & 0.71 &  $0.61 \pm 0.07 $ &  0.32 &  $0.09 \pm 0.10 $   \\
\bottomrule
\end{tabular}
\end{table}

As with the VGG16 + $K$-Means approach we wish to further investigate the clustering results. Taking the best performing MIXAE model on filtered and raw data we tabulate the clusters against their labels. These tables are present in Figs.~\ref{fig:mixae_confmat_filtered} and \ref{fig:mixae_confmat_raw} for filtered and raw data, respectively.

Applied to raw data the MIXAE captures a proton cluster in a similar vein to the $k$-means approach. The MIXAE forms a proton-majority cluster, but with a significant portion of the more noisy proton events being clustered with the other-class. Additionally, we do not observe the carbon events being separated from either the amorphous noise events or from the proton cluster. 

On filtered data the highest performing MIXAE model achieves strong separation of the other-class, but curiously creates two proton-majority clusters.

The most striking result we present in this work is the success of the $k$-means approach. As noted by \cite{Aggarwal}, distance measures become less informative in higher dimensional spaces, but the $k$-means algorithm clusters events well in our fairly high-dimensional VGG latent spaces.
Another surprise is the stability of the $k$-means algorithm. We attribute this stability to the quality of the VGG16 latent space in creating class-separating sub-spaces. While the separations are not perfect, the stability and quality of the proton track identification create solid empirical grounding for applying this approach to other active target experiments.

It is also interesting to compare and contrast the clustering results from the MIXAE model with those of the VGG16 with the  $k$-means approach. In particular, the discrepancy in stability is worth noting. While the top performing MIXAE runs outperform the $k$-means approach, its reliability suffers. However, the high performance achieved indicates that this may represent a valuable potential research path into more tailored models for unsupervised track identification.

\subsection{Alternative approaches}

In addition to the results presented in this section, we performed clustering with a number of different algorithms included in the \texttt{scikit-learn} package of \citet{Pedregosa2011}. None of them provided any notable differences from the $k$-means results or were significantly worse. Notably, the DBSCAN algorithm \cite{Ester96adensity-based,Bergstra2012} failed to provide any useful clustering results. We find this important as one of the significant drawbacks of $k$-means, and the deep clustering algorithm presented in section \ref{sec:mixae}, is that they depend  on pre-determining the number of clusters. This is not the case for DBSCAN. 

Additionally, we considered the deep convolutional embedded clustering (DCEC) approach by \citet{Guo2017} as well as the MIXAE method introduced by \citet{Zhang}. While we were able to reproduce the authors' results on their data, the DCEC algorithm proved unable to cluster AT-TPC data in our implementation. Further details on these experiments are presented in the thesis by \citet{SolliThesis}.

However, this provides valuable insight as the seeds of the clusters are constructed by a $k$-means algorithm. This insight contrasts with our positive results from applying a pre-trained classification model with $k$-means and highlights potentially significant differences in models trained on a supervised or unsupervised objective for clustering tasks in nuclear physics. 

\section{Conclusions and Perspectives}\label{sec:conclusion}
The purpose of this study has been to explore the application of unsupervised learning algorithms to event identification from an active target detector. The necessity to identify events from raw data prior to full processing is becoming a major issue in the data analysis of detectors with complex responses such as the AT-TPC.
As shown by both avenues explored in this work, it is clear that there is significant potential to eventually achieve event classification using fully automated unsupervised methods.

In particular, the ability of the $k$-means algorithm in picking out clear proton clusters from the VGG16 latent space lends itself well to an exploratory phase of analysis, where clusters of events corresponding to different reaction channels could be later identified by the experimenter.
Another interesting facet of the $k$-means clustering is its consistent performance. As shown in Table \ref{tab:clstr_vgg10}, the variance is zero for the performance metrics. This result indicates that the clusters are very clearly defined in the VGG16 latent space. However, as can be seen from the non-proton clusters in Figs.~\ref{fig:clster_confmati_filt} and \ref{fig:clster_confmati_raw},  this does not necessarily imply that the physical signals are correspondingly clear.  Furthermore, the unsupervised metric that decides which of the $M$ $k$-means initializations perform the best does not necessarily coincide perfectly with separating the event classes. This is evidenced by the highest performing model measured on labelled data (Top 1) for the raw data showing up in Table \ref{tab:clstr_vgg1}, and being filtered out from Table \ref{tab:clstr_vgg10}.

A caveat to the $k$-means method is that the number of clusters has to be specified in advance. Each experiment then has to be considered in light of possible reaction channels to determine a sensible number of clusters for this approach. 

The same caveat is present in the MIXAE implementation. While it shows better optimal performance than the $k$-means method, some inconsistencies were observed that were notably not evident from the unsupervised training-objectives of the model. These two factors currently conspire to limit its immediate applicability, and more developments are needed for this approach. 

In summary, our study shows that unsupervised track classification with an implementation of the VGG16 and the  $k$-means approach is a viable solution. For future work, it is worth investigating whether the adaptation of  models like the MIXAE algorithm will allow better performance at no significant cost to consistency.  
The two examples of unsupervised machine learning methods studied in this work are a first encouraging step towards automated selection of similar events that could  greatly reduce the resource cost of analysis. Selection and classification algorithms have the potential to eventually boost the efficiency of the experiment by allowing post-trigger decisions based on such algorithm implemented in hardware. 
\subsection*{Acknowledgements}
MHJ’s work is supported by the U.S. Department of Energy, Office of Science, office of Nuclear Physics under grant No. DE-SC0021152 and U.S. National Science Foundation Grants No. PHY-1404159 and PHY-2013047. DB's work is supported by the U.S. Department of Energy, Office of Science, Office of Nuclear Physics, under Grant No. DE-SC0020451. This material is based upon work supported by the National Science Foundation under Grant No. PHY-2012865  (MPK). This project has also received support from the INTPART project of the Research Council of Norway (Grant No. 288125) and the Davidson Research Initiative. 

\appendix  

\section{VGG16}\label{app:vgg}

The VGGNet models are a family of high-performing image classification, and object localization networks. In the VGGNet architecture a small kernel size is leveraged to increase expressive power in a very deep convolutional network. A tabulated view of the VGG models can be seen in \cite{Simonyan2014}. 

The choice of the kernel size is based on the fact that a stacked $3 \times 3$ kernel is equivalent to larger kernels in terms of the receptive field of the output. Three $3 \times 3$ kernels with stride $1$ have a $7 \times 7$ receptive field, but the larger kernel has $81\%$ more parameters and only one non-linearity \cite{Simonyan2014}. Stacking the smaller kernels then contributes to a lower computational cost. Additionally, there is a regularizing effect from the lowered number of parameters and increased explanatory power from the additional non-linearities.

VGGnet models are distributed freely with weights trained on the ImageNet \cite{deng2009imagenet} image classification task. For the results in section \ref{sec:kmeans_results} we used a VGG16 model pre trained on ImageNet data.

\section{MIXAE hyper-parameter tuning}\label{ax: mixae}

In the MIXAE algorithm the hyper-parameters to adjust are all the ordinary parameters associated with a neural network. We chose to base our neural network parameter choices on the VGG16 architecture. 
The parameters chosen for the autoencoders are listed in full in Table \ref{tab:mixe_ae_hyperparams}.

In addition to those parameters we have the weighting of the loss terms: $\theta$, $\alpha$ and $\gamma$. These weighting parameters are attached to the reconstruction loss, sample entropy and batch-wise entropy respectively \cite{Zhang}. 
We focused on the tuning of the clustering hyper-parameters, and defined the autoencoder hyper-parameters to be a shallow $3\times3$ convolutional network as detailed in the previous paragraph.  

To train the MIXAE clustering algorithm, we use a large simulated data set with $M=80000$ points, evenly distributed between proton- and carbon-events. The algorithm is trained on a subset of $60000$ of these samples, and we track performance on the remaining $20000$ events. On real data the algorithm is trained on an unlabelled set of data, and evaluated on a labelled subset\footnote{see Table \ref{tab:data sets} for details}. Since there are then only three remaining hyperparameters we choose to perform a coarse grid-search for the optimal configuration. Finally, for the best parameters we re-ran the algorithm $N=10$ times to investigate the stability of the algorithm.

\begin{table}
\centering
\caption{Hyperparameter grid for the MIXAE loss weighting terms. The grid is given as exponents for logarithmic scales.}\label{tab:mixae_loss_weights}
\begin{tabular}{lll}
\toprule
Parameter & Grid & Scale \\
\midrule 
$\theta$ & $[-1,\, 5]$ & Logarithmic \\
$\alpha$ & $[-5,\, -1]$ & Logarithmic \\
$\gamma$ & $[3,\, 5]$ & Logarithmic
\end{tabular}
\end{table}

The grids selected for the search are listed in Table \ref{tab:mixae_loss_weights}. The search yielded an optimal configuration with 

\begin{align}
\theta = 10^{-1}, \\
\alpha = 10^{-2}, \\
\gamma = 10^5.
\end{align}

For the full data set the MIXAE hyperparameters converge to the same values as for the clean data:

\begin{align}
\theta &= 10^{1}, \\
\alpha &= 10^{-1}, \\
\gamma &= 3.162\times 10^3.
\end{align}

Lastly we supply the configuration used for the individual convolutional autoencoder networks in Table \ref{tab:mixe_ae_hyperparams}

\begin{table}[H]
\renewcommand*{\arraystretch}{0.5}
\centering
\caption{Hyperparameters selected for the autoencoder components of the MIXAE algorithm}\label{tab:mixe_ae_hyperparams}
\setlength{\extrarowheight}{15pt}
\hspace*{-0.5in}
\begin{tabular}{ll}
\toprule
Hyperparameter & Value \\
\midrule
\multicolumn{2}{l}{Convolutional parameters: } \\
\midrule
Number of layers & $4$ \\
Kernels & $[3,\,3,\,3,\,3]$\\
Strides & $[2,\,2,\,2,\,2]$ \\
Filters & $[64,\, 32, \,16, \,8,]$ \\ 
\midrule
\multicolumn{2}{l}{Network parameters: } \\
\midrule
Activation & LReLu \\
Latent dimension & 20  \\
Batchnorm & False \\
\midrule
\multicolumn{2}{l}{Optimizer parameters: } \\
\midrule
$\eta$ & $10^{-3}$ \\
$\beta_1$ & $0.9$ \\
$\beta_2$ & $0.99$ \\
\bottomrule
\end{tabular}
\end{table}

\section{Data}

The data used  for the analysis in this work were partitioned as shown in Table \ref{tab:data sets}. 
\begin{table}[hbtp]
\centering
\caption{Descriptions of number of events in the data.}\label{tab:data sets}
\begin{tabular}{lccc}
\toprule
{} & Simulated & Full & Filtered \\
\midrule
Total &  $8000$ & $51891$ & $49169$ \\
Labelled & $2400$ & $1774$ &  $1582$ \\ 
\bottomrule
\end{tabular}
\end{table}

\section{Clustering confusion tables}

To elucidate the results presented in Tables \ref{tab:clstr_mixae} and \ref{tab:clstr_vgg1} we computed clustering confusion-matrices. These matrices show a more detailed picture of intermingled classes in a clustering or classification task where ground truth labels are available. In the figures below we tabulate the clusters, as predicted by the algorithm, along the x-axis. Each cluster is decomposed in its ground-truth members along the y-axis.

In this view, a perfect clustering algorithm will produce a confusion matrix which only has nonzero elements in the primary diagonal under free permutation of its columns.

\begin{figure}
\centering
	\includegraphics[width=\textwidth]{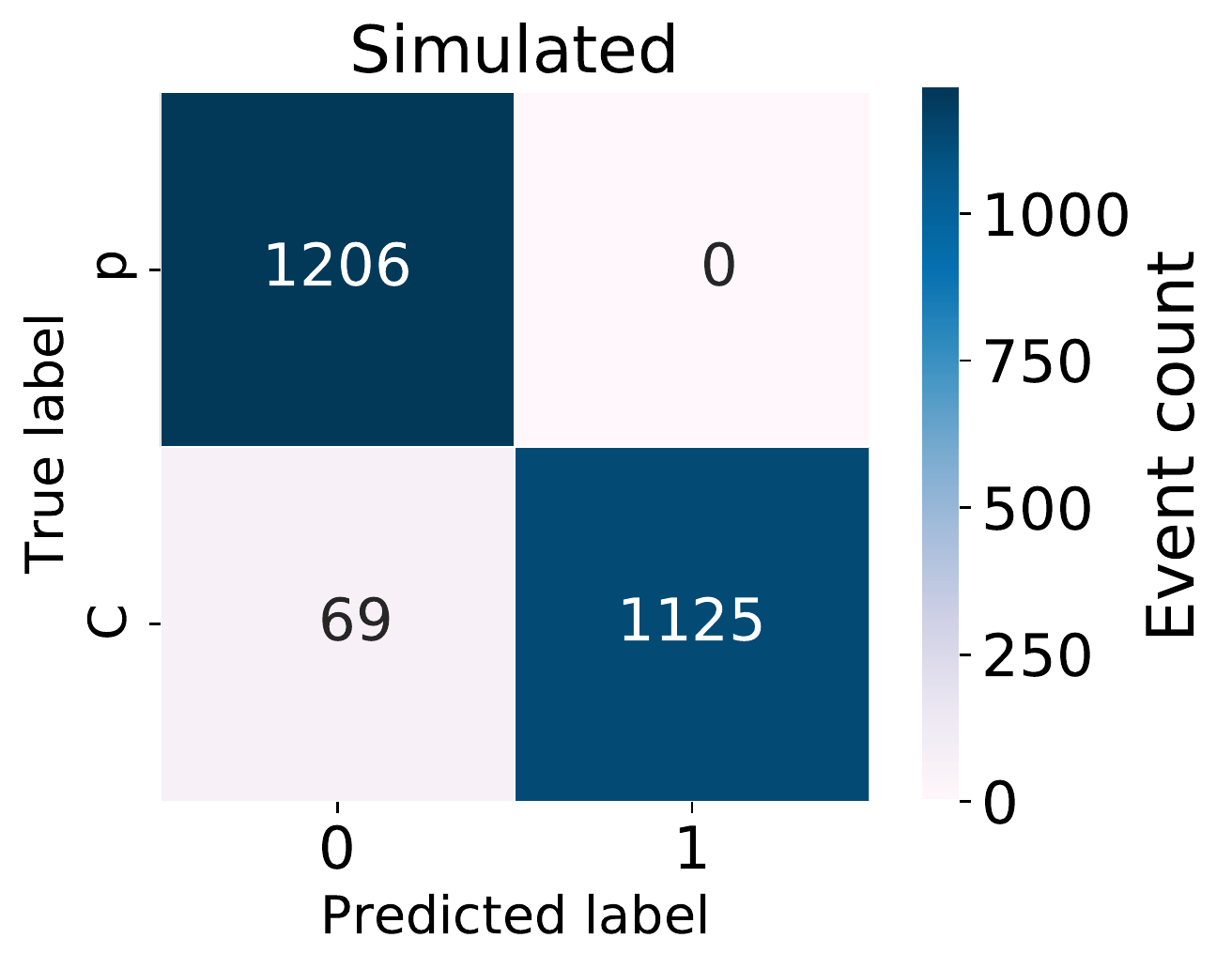} 
	\caption[Pre-trained network - confusion matrices]{Confusion matrix for the $k$-means clustering of simulated AT-TPC events. The true labels indicate samples belonging to the p (proton), or the carbon (C) class. We observe very high quality separation between the proton and carbon classes.}\label{fig:clster_confmat_sim}
\end{figure}

\begin{figure}
\centering
	\includegraphics[width=\textwidth]{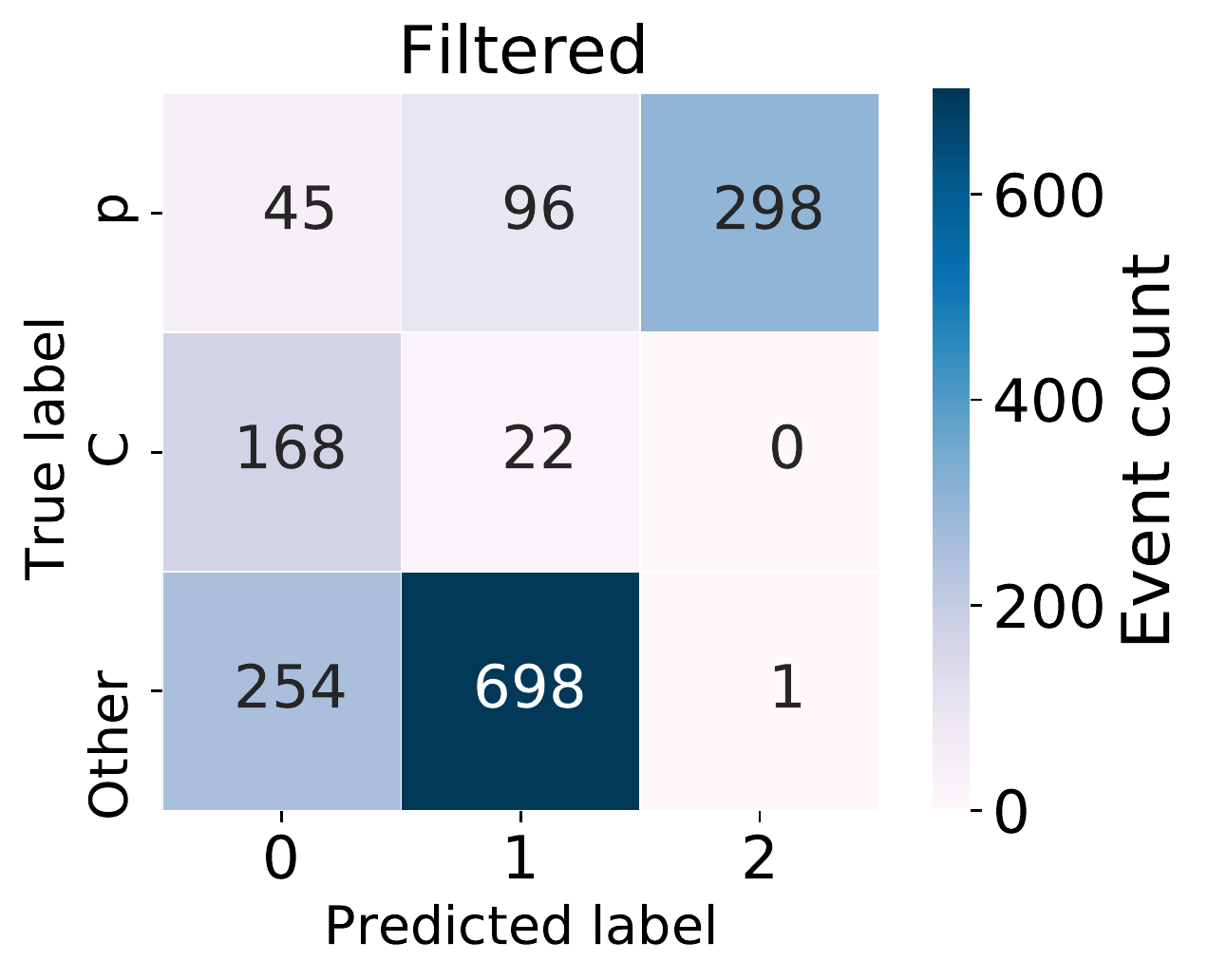}
	\caption[Pre-trained network - confusion matrices]{Confusion matrix for the $k$-means clustering of filtered AT-TPC events. The true labels indicate samples belonging to the p (proton), carbon (C), or other classes. Each column denotes a cluster, with each cell in the column denoting the count of that rows' class in the cluster. We observe that cluster 2 is a high quality proton event cluster }\label{fig:clster_confmati_filt}
\end{figure}

\begin{figure}
	\includegraphics[width=\textwidth]{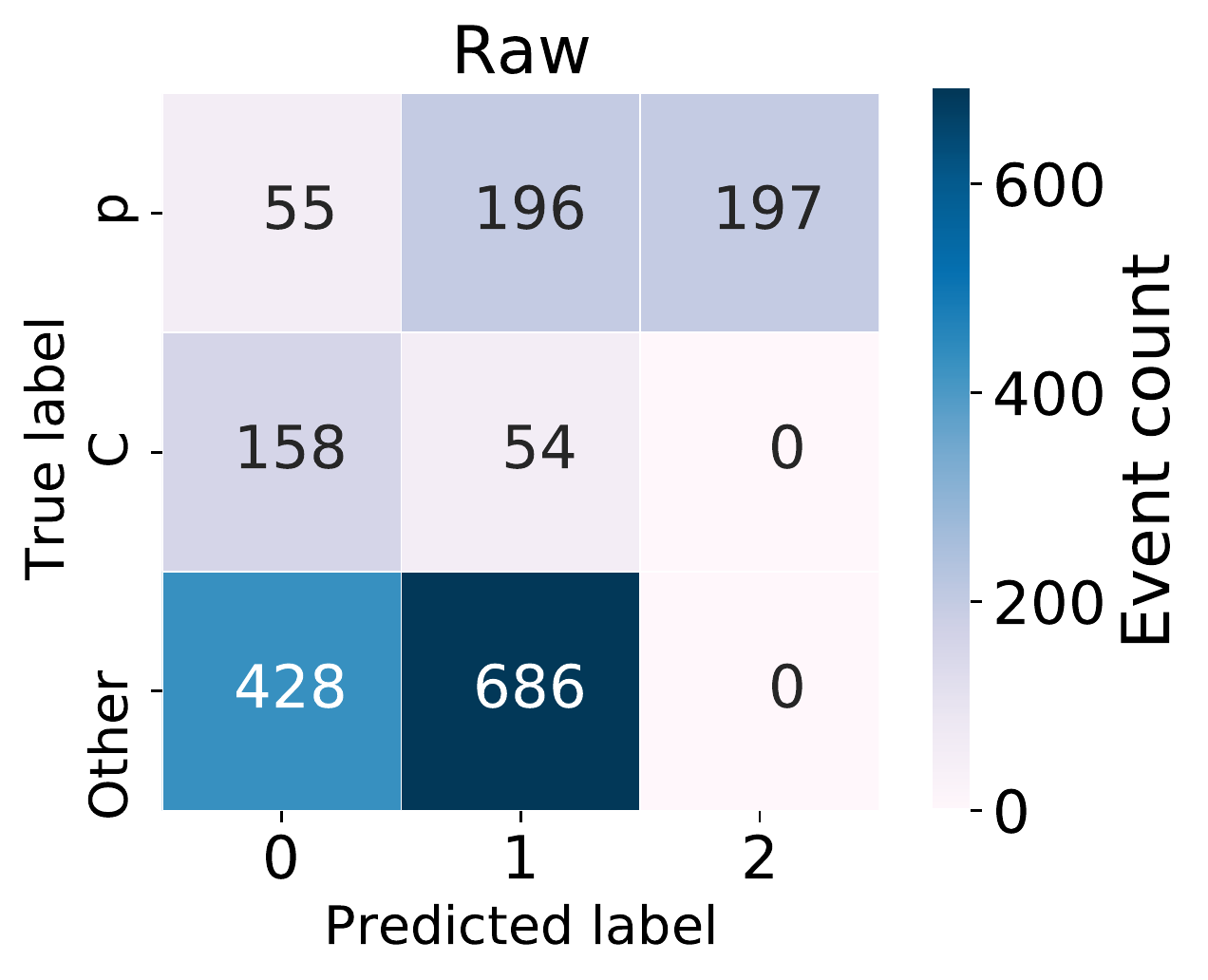}
\caption[Pre-trained network - confusion matrices]{Confusion matrix for the $k$-means clustering of raw AT-TPC events. The true labels indicate samples belonging to the p (proton), carbon (C), or other classes. Each column denotes a cluster, with each cell in the column denoting the count of that rows' class in the cluster. We observe that cluster 2 is a high quality proton event cluster }\label{fig:clster_confmati_raw}
\end{figure}

\begin{figure}
\centering
	\includegraphics[width=\textwidth]{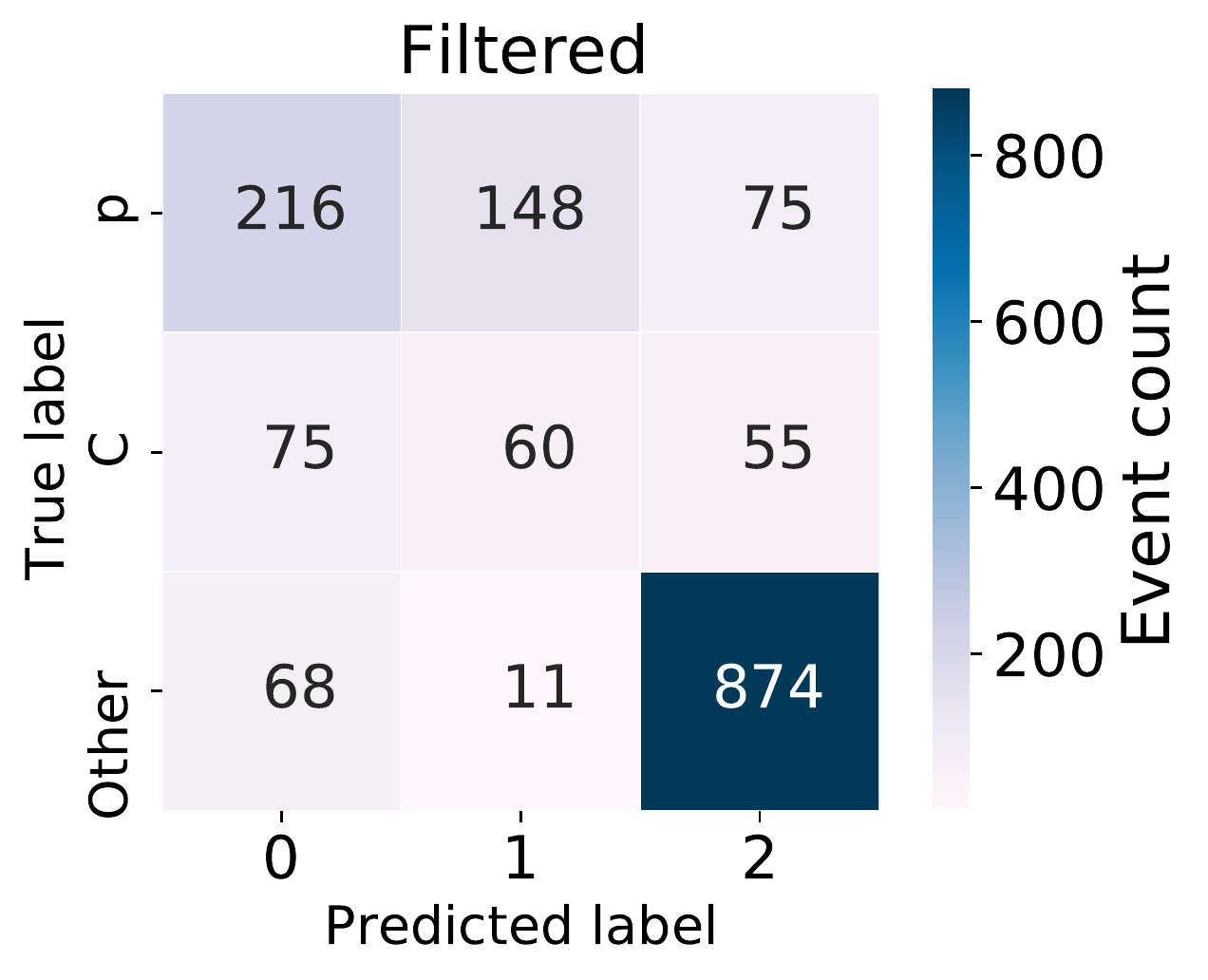}
\caption[MIXAE - confusion matrices]{Confusion matrix for the MIXAE clustering algorithm on filtered AT-TPC events. The true labels indicate samples belonging to the \texttt{p} (proton), \texttt{C} (Carbon), or \texttt{other} classes. We observe that the algorithm forms two proton-majority clusters, and one clearly defined cluster of the other events. }\label{fig:mixae_confmat_filtered}
\end{figure}

\begin{figure}
\centering
	\includegraphics[width=\textwidth]{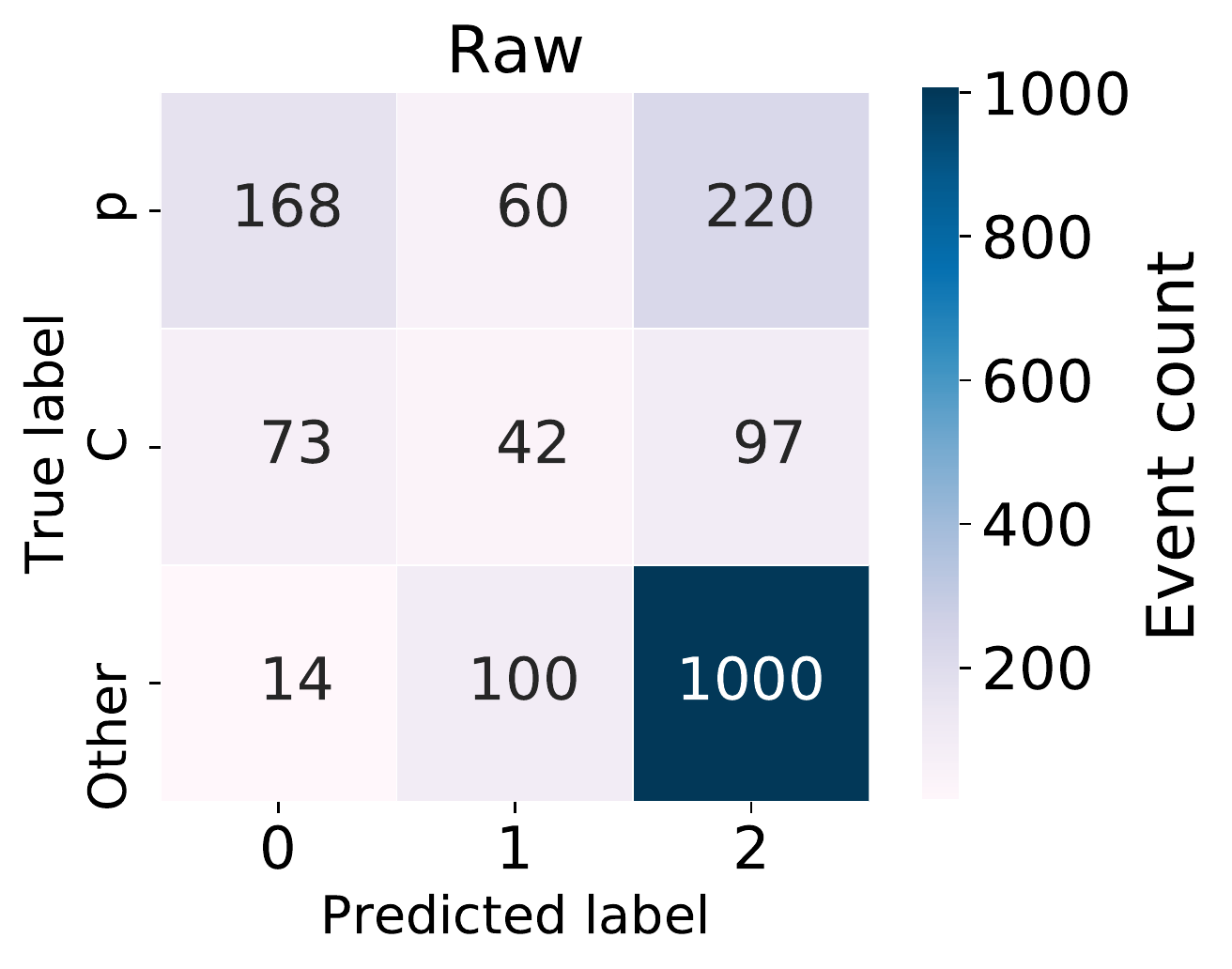}
\caption[MIXAE - confusion matrices]{Confusion matrices for the MIXAE clustering algorithm on raw AT-TPC events. The true labels indicate samples belonging to the p (proton), carbon (C), or \texttt{other} classes. We observe that the algorithm correctly captures a majority proton-event cluster in cluster 0. However, in contrast with the $k$-means approach this cluster is contaminated to some extent with both carbon events and other events. }\label{fig:mixae_confmat_raw}
\end{figure}

\bibliography{article}
\end{document}